\documentclass[smallextended]{svjour3}      
 \usepackage{multirow}
 \usepackage{booktabs}
 \usepackage{subfigure}
\usepackage{amssymb}
\usepackage{times}
\usepackage{caption}
\usepackage{epsfig}
\usepackage{graphicx}
\usepackage{amsmath}
\usepackage{amssymb}
\usepackage{multirow}
\usepackage{placeins}
\usepackage{blindtext}
\usepackage{makecell}
\usepackage{colortbl}
\usepackage{subfigure}
\usepackage{overpic}
\usepackage{bm}
\usepackage{array}
\usepackage{tabulary}
\usepackage{microtype}
\usepackage{booktabs}
\usepackage[table]{xcolor}
\usepackage{soul}
\usepackage[numbers,sort]{natbib}

\begin{document}

\title{SGM-Net: Semantic Guided Matting Net}


\author{Qing Song \and Wenfeng Sun \and Donghan Yang \and Mengjie Hu \and Chun Liu}

\institute{
	Qing Song \at
	Pattern Recognition and Intelligence Vision Lab, Beijing University of Posts and Telecommunications \\
	\email{priv@bupt.edu.cn}  
	\and
	Wenfeng Sun \at
	Pattern Recognition and Intelligence Vision Lab, Beijing University of Posts and Telecommunications \\
	\email{swf980126@bupt.edu.cn}
	\and
	Donghan Yang \at
	Pattern Recognition and Intelligence Vision Lab, Beijing University of Posts and Telecommunications \\
	\email{yangdonghan@bupt.edu.cn}
	\and
	Mengjie Hu \at
	Pattern Recognition and Intelligence Vision Lab, Beijing University of Posts and Telecommunications \\
	\email{mengjie.hu@bupt.edu.cn}  
	\and
	Chun Liu \at
	Pattern Recognition and Intelligence Vision Lab, Beijing University of Posts and Telecommunications \\
	\email{chun.liu@bupt.edu.cn}  
	\and
}

\date{Received: date / Accepted: date}

\maketitle

\begin{abstract}
Human matting refers to extracting human parts from natural images with high quality, including human detail information such as hair, glasses, hat, etc. This technology plays an essential role in image synthesis and visual effects in the film industry. When the green screen is not available, the existing human matting methods need the help of additional inputs (such as trimap, background image, etc.), or the model with high computational cost and complex network structure, which brings great difficulties to the application of human matting in practice. To alleviate such problems, most existing methods (such as MODNet) use multi-branches to pave the way for matting through segmentation, but these methods do not make full use of the image features and only utilize the prediction results of the network as guidance information. Therefore, we propose a module to generate foreground probability map and add it to MODNet to obtain Semantic Guided Matting Net (SGM-Net). Under the condition of only one image, we can realize the human matting task. We verify our method on the P3M-10k dataset. Compared with the benchmark, our method has significantly improved in various evaluation indicators.
\end{abstract}


\maketitle

\section{Introduction}\label{sec1}

Semantic segmentation directly identifies the object category of each pixel, which belongs to rough semantics and is easy to blur the structural details. Human parsing is a fine-grained semantic segmentation task for human images, aiming to identify the components of human images at the pixel level\cite{yang2022part, yang2020renovating, he2017real}. Although human parsing enhances the processing of structure information, it is still essentially pixel-level coarse extraction. Different from them, image matting is the technique of extracting the foreground from a natural image by calculating its color and transparency. It can be used for background replacement, image synthesis, and visual effects, which has a wide application prospect in the film industry\cite{wang2008image,levin2007closed}.

Image matting needs to calculate the transparency of each pixel, which is more refined than human parsing. Specifically, for the input image $\emph{I} \in \mathbb{R}^{H \times W \times 3}$, the extinction formula is decomposed into foreground $\emph{F} \in \mathbb{R}^{H \times W \times 3}$, background $\emph{B} \in \mathbb{R}^{H \times W \times 3}$, and alpha matte $\alpha \in \mathbb{R}^{H \times W} $ with a linear mixing assumption:
\begin{align}
    \label{eq_1}I = \alpha F + (1 - \alpha)B
\end{align}
where for color images, there are 7 unknown variables in the above expression and only 3 known variables, and thus, this decomposition is seriously limited\cite{levin2007closed}.

Most of the existing matting methods need additional pictures as auxiliary inputs, such as additional background images \cite{2020Real,BGMatting}, and pre-defined trimaps \cite{wang2008image,attention3,dim}. However, taking additional background image as input requires that the two pictures must be aligned. Besides, the cost of pre-defined trimap is too high for us. Therefore, some latest work attempts to study the matting problem in a trimap-free setting. These researches include two directions: one is to study the alternatives to the trimap guidance and ease the requirements for manual input \cite{gupta2016automatic,liu2020boosting,hsieh2013automatic,yu2021mask}. For example, \cite{gupta2016automatic,hsieh2013automatic} proposed technologies of automatic generation of trimap, while \cite{yu2021mask} proposed the progressive refinement network (PRN), which has good robustness to various types of masks as guidance. Another line of works try to get rid of any external guidance, hoping that the matting model can capture both semantics and details through end-to-end training on large-scale datasets \cite{qiao2020attention,zhang2019late,modnet,lin2022robust}, and achieve the same level of video matting as video segmentation \cite{li2022locality}.

Compared with common objects, portraits have more abundant and complex details, such as hair, glasses, jewelry, etc. These factors make human matting more challenging. Therefore, in this work, we focus on the human matting and design a module to generate foreground probability map using the features in the segmentation network. This foreground probability map has the same size as the input image, which can be used as the guidance information to assist in the detail extraction of the human contour. By adding it to MODNet, we get a new human matting network SGM-Net, which uses a single RGB image as input to complete four tasks: extracting human contour, generating foreground probability map, predicting detail information and fusing information, so as to predict accurate alpha matte. 

We conduct a large number of experiments using P3M-10k dataset\cite{p3m-10k} to evaluate the effectiveness of our method. Under the commonly used matting performance indicators, our method has excellent performance. For details, please refer to Sec.\ref{sec4}. In addition, we capture a large number of natural human images from the Internet, which proves that our learning model can be extended to real-images.
\begin{figure}[htbp]
	\centering
	\subfigure[Input]
	{
		\begin{minipage}[b]{.28\linewidth}
			\centering
			\includegraphics[scale=0.4]{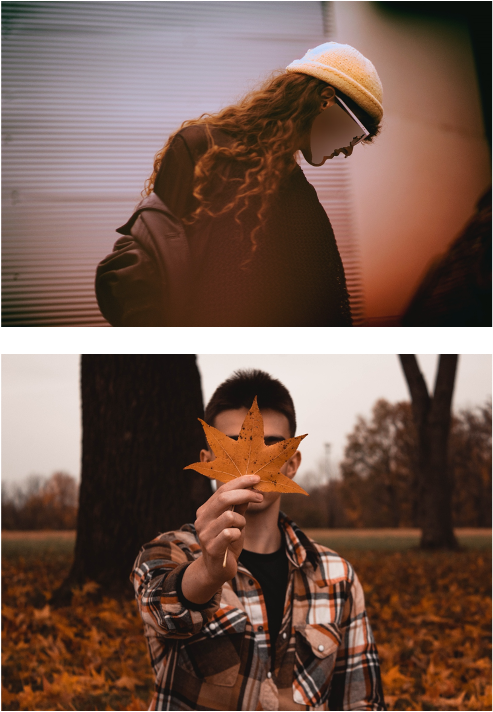} 
		\end{minipage}
	}
	\subfigure[MODNet\cite{modnet}]
	{
		\begin{minipage}[b]{.28\linewidth}
			\centering
			\includegraphics[scale=0.4]{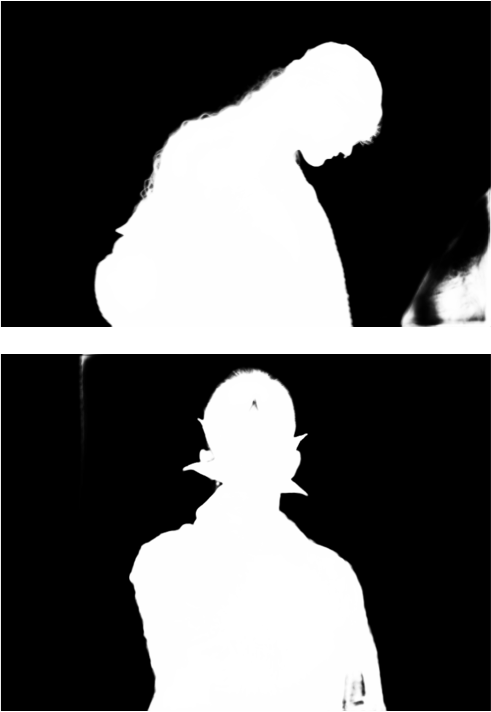} 
		\end{minipage}
	}
	\subfigure[Ours]
	{
		\begin{minipage}[b]{.28\linewidth}
			\centering
			\includegraphics[scale=0.4]{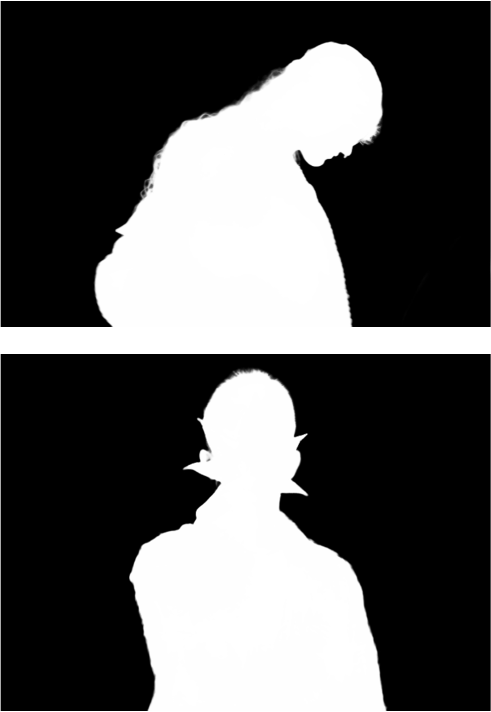} 
		\end{minipage}
	}
	\caption{Alpha matte results by MODNet and our SGM-Net from RGB input.}\label{fig1}
\end{figure}

\section{Related Works}\label{sec2}

Image matting is to extract the desired target foreground from a given target image. Unlike binary mask output by image segmentation \cite{minaee2021image, yu2018semantic} and human parsing \cite{yang2019parsing, yang2021quality}, image/human matting needs to be given an alpha matte with accurate prediction probability of foreground for each pixel, which is represented by $\alpha$ in Eq.\ref{eq_1}. In this section, we will review the image matting methods related to our work.

\subsection{Traditional Methods}\label{ssec2.1}

Traditional image matting methods mostly predict alpha matte by sampling, propagation or using low-level features such as color \cite{color1,color3}. The sampling-based methods \cite{caiyang1,caiyang2,caiyang3} estimate foreground/background color statistics by sampling the pixels in the definite foreground/background regions, so as to solve the alpha matte of the unknown region. The propagation-based method \cite{chuanbo1,chuanbo2,chuanbo3,chuanbo4,chuanbo5} estimate alpha matte by propagating the alpha value of the foreground/background pixels to the unknown region. However, the application effect of these methods in complex scenes is not ideal. 

\subsection{Trimap-based Methods}\label{ssec2.2}

With the great progress of deep learning and the rise of computer vision technology, many methods based on convolutional neural network (CNN) are proposed and used in general image matting, which significantly improves the matting results. Cho et al. \cite{cho} and Shen et al. \cite{shen} introduced convolutional neural network into the traditional algorithm to reconstruct the alpha matte. Xu et al. \cite{dim} proposed an automatic encoder architecture, which takes RGB image and trimap as input and uses pure encoder-decoder network to directly predict alpha matte to achieve the state-of-the-art results. The trimap is a mask containing three regions: absolute foreground ($\alpha=1$), absolute background ($\alpha=0$) and unknown region ($\alpha=0.5$). In this way, the matting algorithm only needs the prior information of two absolute regions to predict the masking probability of each pixel in the unknown region. Later,  Lutz et al. \cite{alphagan} introduced a generative adversarial framework to improve the results. Tang et al. \cite{tang2019} proposed to combine the sampling-based method with deep learning. Hou et al. \cite{hou2019context} proposed a two-encoder two-decoder structure for simultaneous estimation of foreground and alpha. With the development of the attention mechanism, \cite{yang2021attacks, yang2018attention} introduced the attention mechanism into the human parsing task, which greatly improved the accuracy. Furthermore, research \cite{attention1,attention2} argued that the attention mechanism can effectively improve matting performance. \cite{attention3} further improved the performance by introducing the contextual attention module.

\subsection{Trimap-free Methods}\label{ssec2.3}
Semantic estimation is needed to locate the approximate foreground before predicting the fine alpha matte, thus, it is difficult to denoise the image without the assistance of trimap.

Currently, trimap-free methods mostly focus on specific types of foreground objects, such as human and animals \cite{bridging}. \cite{zhang2019late} proposed a framework composed of a segmentation network and a fusion network, where the input is only a single RGB image. Then, \cite{liu2020boosting} introduced a trimap-free framework, which consisting of mask prediction network, quality unification network and matting refinement network. Similarly, \cite{modnet} proposed a framework composed of semantic estimation network, detail prediction network and semantic-detail fusion network, and introduced the attention mechanism --- SE module \cite{se}. As the representatives of another direction of trimap-free matting research, Chen et al. \cite{shm} realized foreground detail extraction by combining trimap generation network (T-Net) and matting network (M-Net). \cite{BGMatting} introduced a framework that takes the background images along with other potential prior information (such as segmentation mask, motion cue, etc.) as additional inputs. \cite{yu2021mask} designed Progressive Refinement Network (PRN) to provide self-guidance for learning and progressively refines the uncertain matting areas by decoding. The model can use various types of mask guidance (such as trimap, rough binary segmentation mask or low-quality soft alpha matte) to obtain high-quality matting results and weaken the dependence of the model on trimap.

Our method directly learns the semantic information of the given images for the human foreground, generates foreground probability map while roughly extracting the human contour, extracts the fine details according to the foreground probability map and some semantic information, these are finally fused to generate accurate alpha mattes. Based on the original MODNet, a small amount of model operation cost is sacrificed to effectively improve the functioning of human matting.

\section{Method}\label{sec3}
In this section, we will introduce our algorithm with more details.

\subsection{Overview}\label{ssec3.1}
\begin{figure}[h]%
	\centering
	\includegraphics[width=0.9\textwidth]{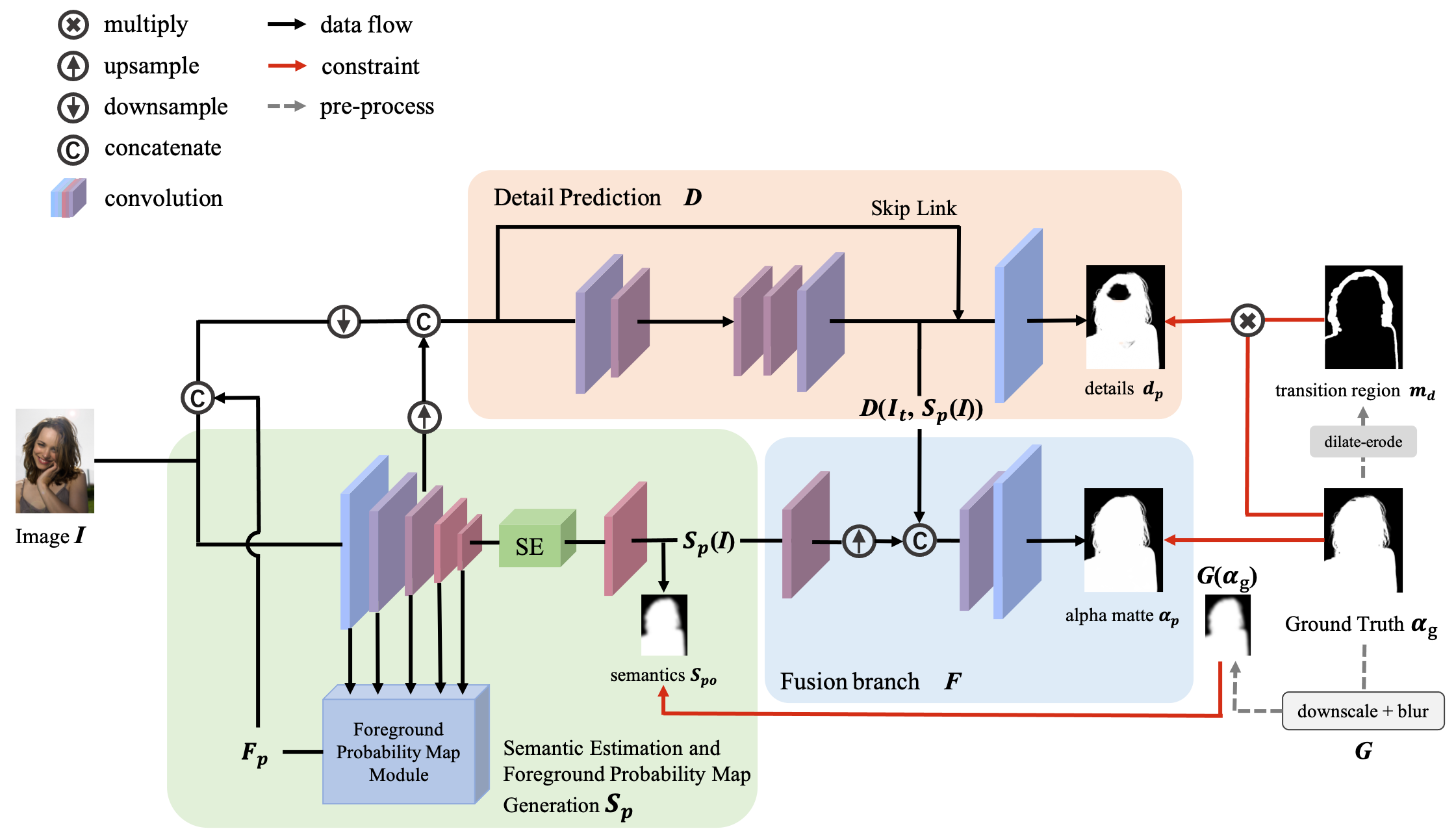}
	\caption{Architecture of SGM-Net. Given an input image \textit{I}, SGM-Net predicts human semantics ${S}_{p}{(I)}$, foreground probability map $F_p$, boundary details $d_p$ and final alpha matte $\alpha_p$. All branches are constrained by specify supervisions generated from the ground truth matte $\alpha_g$.}\label{fig2}
\end{figure}

Our SGM-Net is targeted to extract alpha mattes of specific semantic patterns – human. SGM-Net takes an image (usually 3 channels representing RGB) as the input and directly outputs a 1-channel alpha matte with the identical size of input, no auxiliary information (such as trimap or background image) is required. Fig. \ref{fig2} shows its pipeline. Methods that are based on multiple models \cite{shm,shen,salient} have shown that regarding the trimap-free matting as a combination of trimap prediction (or segmentation) step and trimap-based matting step can obtain better performance, but the large amount of computation brought by multiple models is not as expected.
We decompose the trimap-free matting task into multiple subtasks including semantic estimation, probability prediction, detail prediction and information fusion to expand and optimize this idea. Although our results are not able to surpass the trimap-based methods, they are better than the trimap-free methods based on multiple models.

The objective of SGM-Net is to obtain accurate alpha matte by generating coarse semantic classification information and fine foreground boundaries. As shown in Fig. \ref{fig2}, SGM-Net consists of three branches, which can learn different sub-objectives. Specifically, the low-resolution branch (${S}_{p}$) of SGM-Net uses a segmentation network to estimate human semantics, and in this process, it does pixel-wise classification among foreground and background to generate foreground probability map. Based on it, the high-resolution branch (\emph{D}) (supervised by the transition region ($\alpha \in (0,1)$ in the ground truth matte) concatenates the input image (RGB) and foreground probability map to focus on portrait boundaries. Finally, at the end of the model, the fusion branch (\emph{F}) (supervised by the ground truth matte) integrates rough human semantics and fine human boundary information to predict the final alpha matte. The whole networks are trained jointly in an end-to-end manner. We describe these branches in detail in the following sections.
\begin{figure}[h]%
	\centering
	\includegraphics[width=0.9\textwidth]{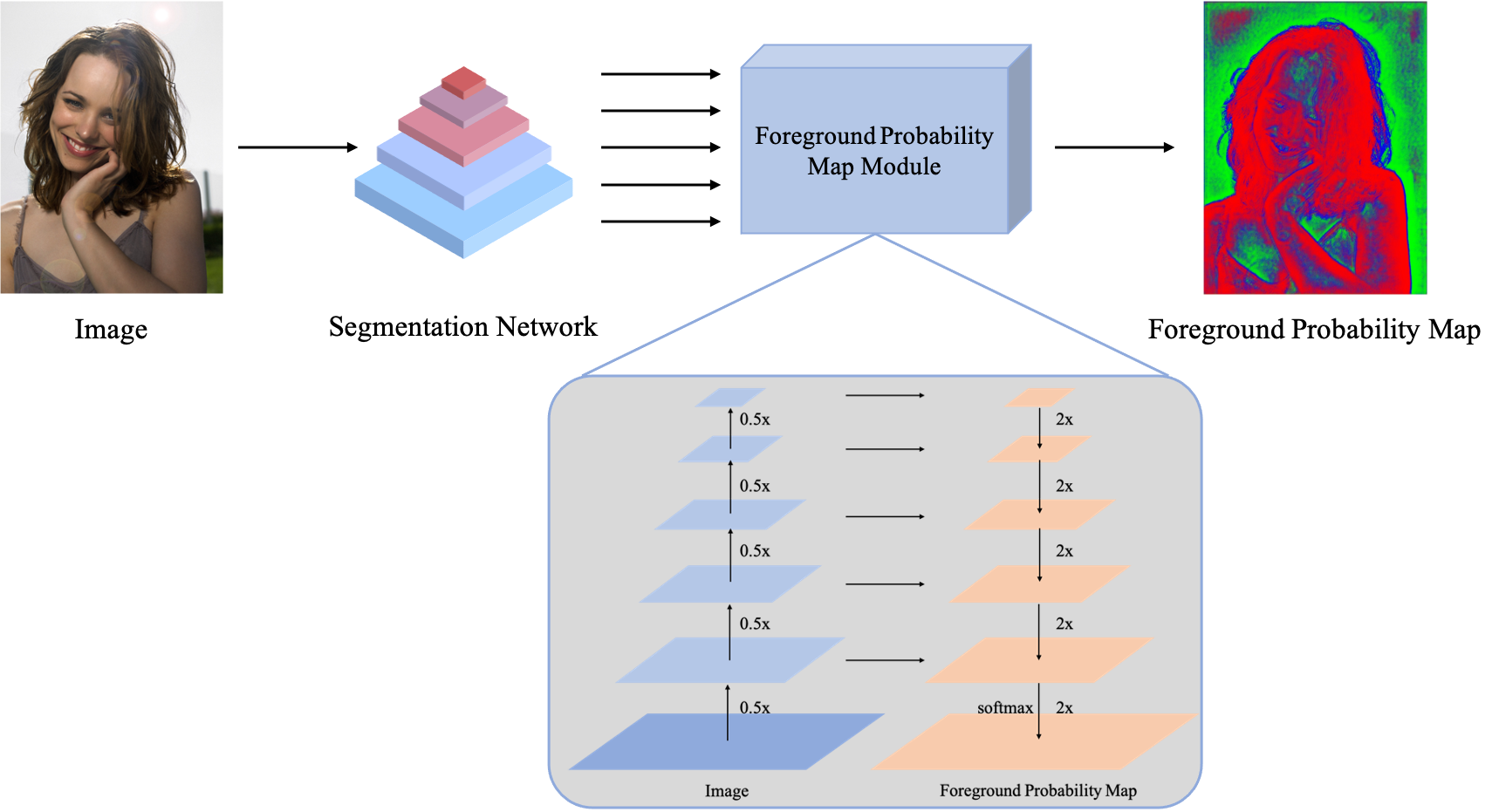}
	\caption{Architecture of Foreground Probability Map Module. Given an image as input, the segmentation network extract features of different scales, input all features into our module, use encoder-decoder to recombine all features, and finally use softmax to obtain the probability that each pixel of the original image is the foreground of portrait. Visualize the probability map, the darker the color, the greater the probability that the pixel is the foreground of the person, and vice versa, it is more likely to be the background.}\label{fig3}
\end{figure}

\subsection{Semantic Estimation and Foreground Probability Map Generation}\label{ssec3.2}

Similar to existing trimap-free methods, the first step of SGM-Net is to locate the human in the input image \emph{I}. we extract the high-level semantics of the image through an encoder (i.e., the low-resolution branch ${S}_{p}$ of SGM-Net), which makes the semantic estimation more efficient. In addition, ${S}_{p}{(I)}$ is helpful for subsequent branches and joint optimization. Experiments in \cite{modnet} show that some channels have more precise semantics than other channels, thus, the channel-wise attention mechanism can encourage using the right information and discourage those that are wrong. Therefore, we continue to use SE-block after ${S}_{p}$ to reweight the channels of ${S}_{p}{(I)}$ by extracting features of different scales. Further, in order to predict the rough semantic mask, we feed ${S}_{p}{(I)}$ into a convolution layer activated by the Sigmoid function to change its channel number to 1. Since the semantic mask is supposed to be smooth, we use L2 loss here, as shown below:
\begin{align}
	\label{eq_2}\mathcal{L}_{s}=\frac{1}{2}\parallel S_{po}-G(\alpha_{g})\parallel_{2}
\end{align}
where \emph{G} stands for Gaussian blur after 16$\times$ downsampling, which removes the detailed structures (such as hair) that are not important to human semantics.

According to Eq.\ref{eq_1}, the output of the matting network can be regarded as the probability prediction ($\alpha \in (0, 1)$) that each pixel in the image is the foreground. Therefore, we extend the semantic segmentation process of this branch, add a probability prediction module that can fuse the feature information in the segmentation network and generate the probability prediction map of the foreground of the human. As shown in the Fig.\ref{fig3}. The output of the module indicates the probability of each pixel in the input as the foreground, which can be used as guidance information to assist the high-resolution branch \emph{D} to complete the detail extraction of the human contour.

\subsection{Detail Prediction}\label{ssec3.3}

We process the unknown area around the foreground human by using a high-resolution branch \emph{D}, which takes the input image \emph{I}, foreground probability map ${F}_{p}$ and low-level features from ${S}_{p}$ as the input. Reusing low-level features can reduce the computational overheads of \emph{D} and improve the speed of the model. We take the concatenation of 3-channel image \emph{I} and the 3-channel foreground probability map ${F}_{p}$ as 6-channel input $I_t$ of \emph{D}, as shown in Figure \ref{fig1}, the input after downsampling is concatenated with the low-level features in ${S}_{p}$ and transmitted to the encoder-decoder. In \emph{D}, we use a skip link to reduce the impact of resolution change on prediction accuracy.

We use the dependency between sub-objectives, that is, foreground probability map and high-level human semantics ${S}_{p}{(I)}$ are a priori for detail prediction. We denote the output of \emph{D} as ${D(I_t,S_p(I))}$, calculate the portrait boundary details from it and learn it through L2 loss. The expression is as follows:
\begin{align}
	\label{eq_3}\mathcal{L}_{d}=m_d\parallel d_p-\alpha_g\parallel_2
\end{align}
where $m_d$ is a binary mask that makes $\mathcal{L}_{d}$ focus on the boundaries of the portrait. It comes from dilation and erosion on $\alpha_{g}$ . The values are 1 if the pixels are inside the unknown region, otherwise, it is 0. Although $d_p$ may contain inaccurate values for the pixels with $m_d=0$, the values in $d_p$ are highly accurate for the pixels with $m_d=1$.

\subsection{Fusion Module}\label{ssec4.4}

For the fusion branch, we use the concise CNN module to concatenate human semantics and boundary details. First, we upsample the output $S_p(I)$ of the semantic branch, match its shape with ${D(I_t,S_p(I))}$. We then concatenate them to predict the final alpha matte $\alpha_p$. The loss function of this process is as shown in Eq.\ref{eq_4}:
\begin{align}\label{eq_4}
	\mathcal{L}_{\alpha}=\parallel\alpha_p-\alpha_g\parallel_1 + \mathcal{L}_{c}
\end{align}
where $\mathcal{L}_{c}$ is the compositional loss from \cite{dim}. It measures the absolute difference between the input image $I$ and the composite image, which is composed of prediction alpha matte $\alpha_p$, the ground truth foreground, and the ground truth background.

Our SGM-Net realizes end-to-end training by adjusting the weights of $\mathcal{L}_{s}$, $\mathcal{L}_{d}$ and $\mathcal{L}_{\alpha}$, as:
\begin{align}
	\label{eq_5}\mathcal{L}=\lambda_s\mathcal{L}_{s}+\lambda_d\mathcal{L}_{d}+\lambda_\alpha\mathcal{L}_{\alpha}
\end{align}
where $\mathcal{L}_{s}$, $\mathcal{L}_{d}$ and $\mathcal{L}_{\alpha}$ are hyper-parameters that balances the three branch loss functions. Follow the settings in MODNet, we set $\lambda_s=\lambda_\alpha=1$ and $\lambda_d=10$.

\section{Experiments}\label{sec4}

In this section, we compare SGM-Net with MODNet and other methods (such as DIM\cite{dim}, AlphaGAN\cite{alphagan}, SHM\cite{shm}, etc.) on the P3M-10k \cite{p3m-10k} face-blurred images to verify the effectiveness of foreground probability map module. We also conduct further ablation experiments to evaluate the performance of SGM-Net in various aspects. Finally, we demonstrate the effectiveness of SGM-Net in adapting to real-world data.

\subsection{Experiments Setting}\label{ssec4.1}
We train all models on the same dataset and adopt the same training strategy, as follows:

\emph{\textbf{Dataset.}} We evaluate all methods on the human matting dataset (P3M-10k face-blurred), which contains 9,421 training images and 500 testing images.

\emph{\textbf{Measurement.}} Five metrics are used to evaluate the quality of predicted alpha matte: sum of absolute differences (SAD), mean squared error (MSE), mean absolute differences (MAD), gradient (Grad) and Connectivity (Conn). Among them, SAD, MSE and MAD measure and evaluate the pixel difference between the prediction and the ground truth alpha matte, while Grad and Conn measure clear details. In calculating all these metrics, we normalized the predicted alpha matte and ground truth to 0 to 1. Furthermore, all metrics are calculated over the entire images instead of only within the unknown regions and averaged by the number of pixels.

\emph{\textbf{Training Stage.}} To make a fair comparison with MODNet, we use the pre-trained weight of ImageNet \cite{imagenet} to initialize the network. We train all networks on  8 NVIDIA Titan XP GPUs (input images are cropped to 512 $\times$ 512) with a batch size of 4. The momentum of the SGD optimizer is set to 0.9 and the weight decay is 4.0e-5. The learning rate is initialized to 0.02, the training lasts 150 epochs, and is multiplied by 0.1 every 50 epochs.

\subsection{Results on P3M-10k}\label{ssec4.2}
\begin{figure}[htbp]
	\centering
	\subfigure[Image]
	{
		\begin{minipage}[b]{.16\linewidth}
			\centering
			\includegraphics[scale=0.4]{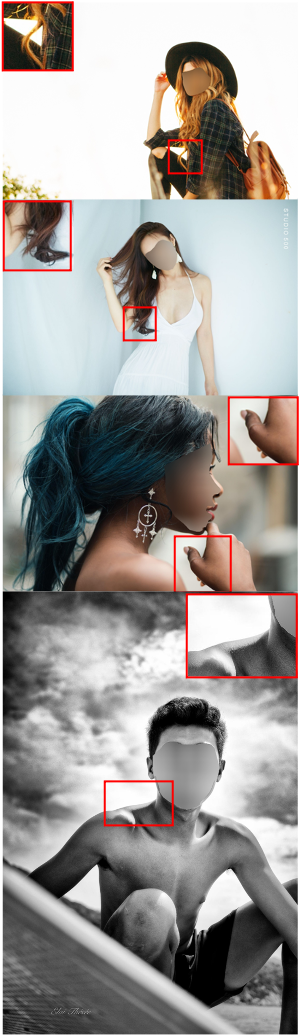} 
		\end{minipage}
	}
	\subfigure[GT]
	{
		\begin{minipage}[b]{.16\linewidth}
			\centering
			\includegraphics[scale=0.4]{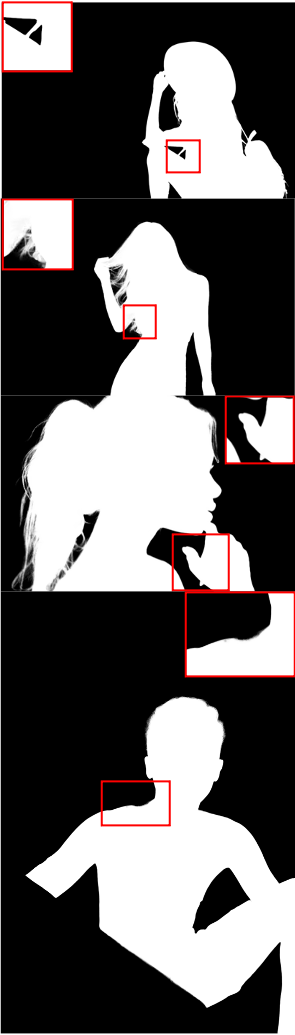} 
		\end{minipage}
	}
	\subfigure[DIM\cite{dim}]
	{
		\begin{minipage}[b]{.16\linewidth}
			\centering
			\includegraphics[scale=0.4]{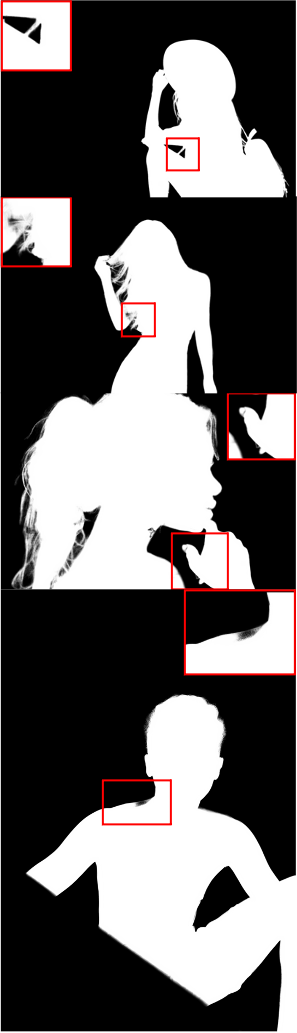} 
		\end{minipage}
	}
	\subfigure[MODNet\cite{modnet}]
	{
		\begin{minipage}[b]{.16\linewidth}
			\centering
			\includegraphics[scale=0.4]{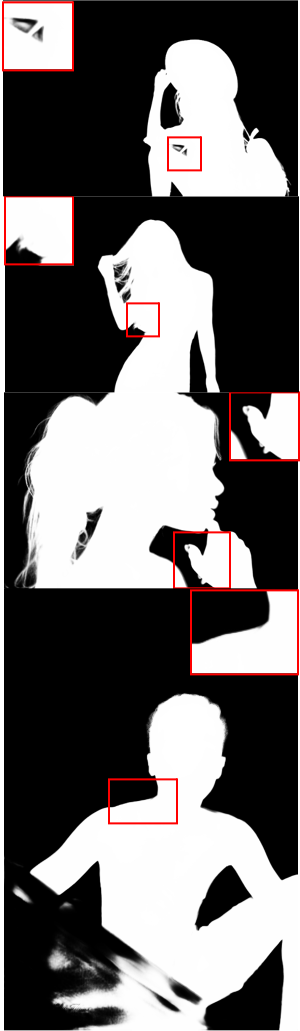} 
		\end{minipage}
	}
	\subfigure[Ours]
	{
		\begin{minipage}[b]{.16\linewidth}
			\centering
			\includegraphics[scale=0.4]{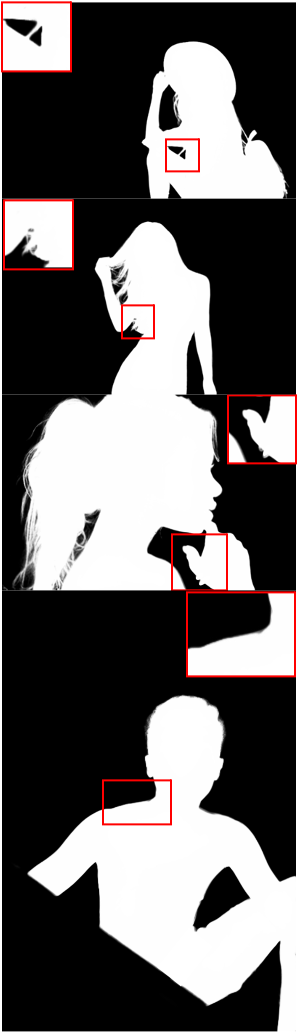} 
		\end{minipage}
	}
	\caption{Subjective Results of Different Methods on P3M-500-P. We test several methods and show the results of some representative methods (DIM\cite{dim}, MODNet\cite{modnet} and ours) on P3M-500-P. Zoom it for the best visualization.}\label{fig4}
\end{figure}

\begin{figure}[htbp]
	\centering
	\subfigure[Image]
	{
		\begin{minipage}[b]{.16\linewidth}
			\centering
			\includegraphics[scale=0.4]{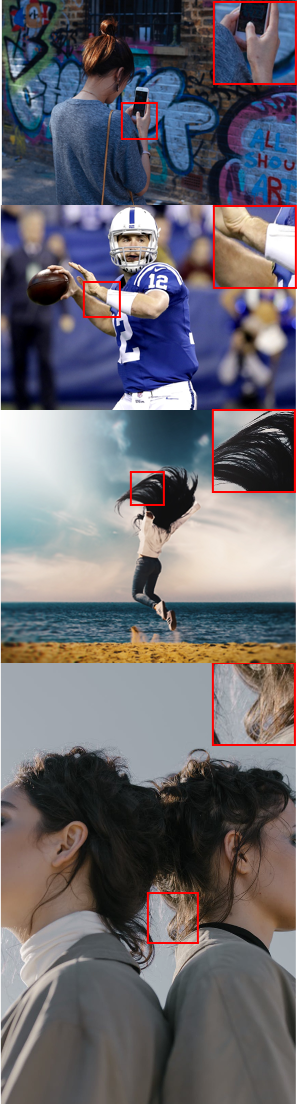} 
		\end{minipage}
	}
	\subfigure[GT]
	{
		\begin{minipage}[b]{.16\linewidth}
			\centering
			\includegraphics[scale=0.4]{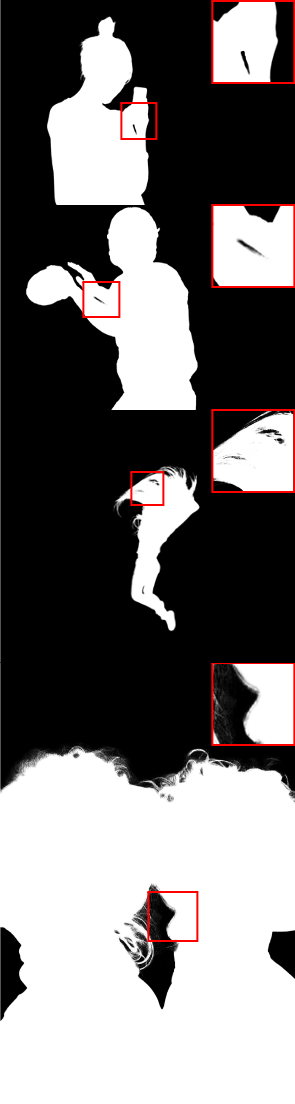} 
		\end{minipage}
	}
	\subfigure[DIM\cite{dim}]
	{
		\begin{minipage}[b]{.16\linewidth}
			\centering
			\includegraphics[scale=0.4]{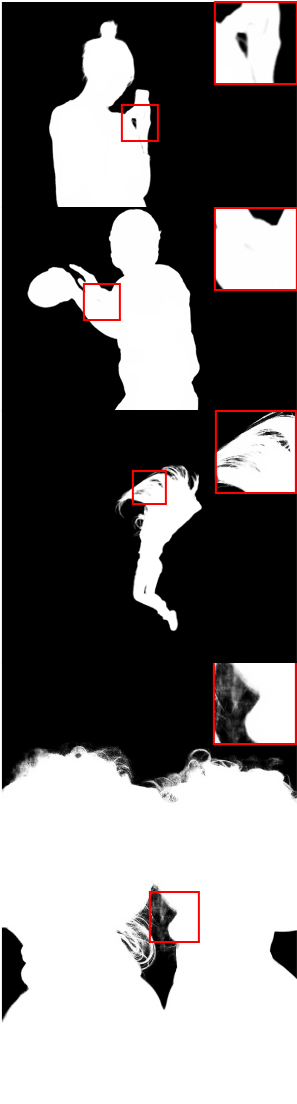} 
		\end{minipage}
	}
	\subfigure[MODNet\cite{modnet}]
	{
		\begin{minipage}[b]{.16\linewidth}
			\centering
			\includegraphics[scale=0.4]{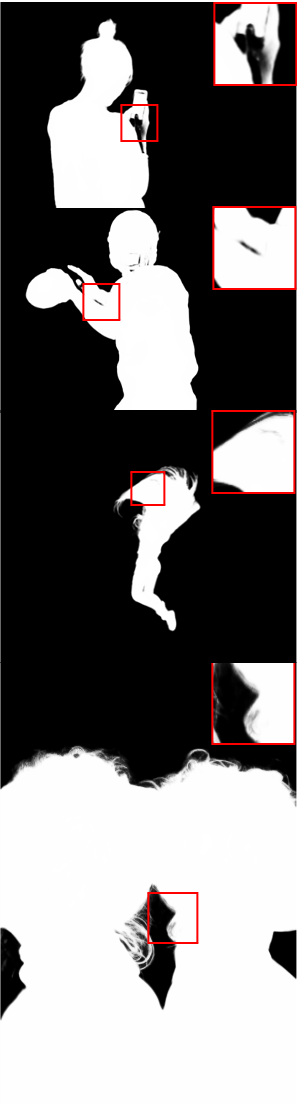} 
		\end{minipage}
	}
	\subfigure[Ours]
	{
		\begin{minipage}[b]{.16\linewidth}
			\centering
			\includegraphics[scale=0.4]{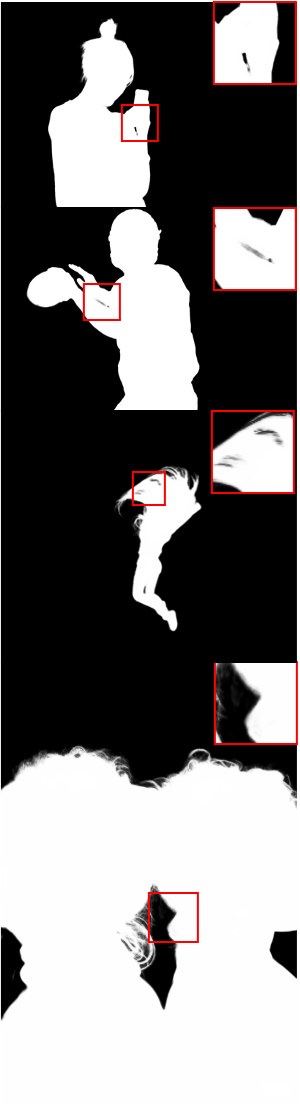} 
		\end{minipage}
	}
	\caption{Subjective Results of Different Methods on P3M-500-NP. We test several methods and show the results of some representative methods (DIM\cite{dim}, MODNet\cite{modnet} and ours) on P3M-500-NP. Zoom it for the best visualization.}\label{fig5}
\end{figure}

In order to evaluate the effectiveness of our proposed method, we use ResNet-34 as the backbone of all trimap-free method and compare SGM-Net with other matting methods. 
As the objective and subjective results of different methods on P3M-10k shown in Table \ref{tab1} and Fig. \ref{fig4}, our method is superior to MODNet and other trimap-free methods in all indicators, and even achieves competitive results with DIM based on trimap. 
The P3M-10k dataset contains two testing datasets, P3M-500-P and P3M-500-NP. P3M-500-P dataset blurs the identifiable faces, while P3M-500-NP does not, see the Image in Fig. \ref{fig4} and Fig. \ref{fig5}. We also perform the same test on P3M-500-NP. The results are shown in Table \ref{tab2} and Fig. \ref{fig5}.
\begin{table}
	\begin{center}
		\scalebox{0.99}{
			\begin{tabular}{c|ccccc}
				\toprule%
				\multicolumn{6}{c}{P3M-500-P} \\
				\midrule
				Method & SAD $\downarrow$  & MSE $\downarrow$  & MAD $\downarrow$  & Grad $\downarrow$  & Conn $\downarrow$ \\
				\midrule
				DIM\cite{dim} & 6.1499 & 0.0011 & 0.0036 & 9.7408 & 6.3404 \\
				AlphaGAN\cite{alphagan} & 6.6239 &	0.0016 & 0.0039 & 18.7622 & 6.8468 \\
				IndexNet\cite{attention2} & 6.5346 &	0.0014 & 0.0038 & 18.4972 &	6.7370 \\
				\midrule
				HATT\cite{qiao2020attention} & 26.9383 & 0.0055 & 0.0156 & 30.0513 & 14.0484 \\
				SHM\cite{shm} & 23.0524 & 0.0098 & 0.0130 & 43.9107 & 9.8490 \\
				MODNet\cite{modnet} & 9.7328 & 0.0031 & 0.0056 & 13.4755 & 9.0085 \\
				Ours & 9.1552 & 0.0027 & 0.0053 & 13.3744 & 8.7354 \\
				\bottomrule
			\end{tabular}
		}
	\end{center}
	\caption{Results of trimap-based methods and trimap-free methods on P3M-500-P testing dataset.}
	\label{tab1}
	\vspace{-.8em}
\end{table}
\begin{table}
	\begin{center}
		\scalebox{0.99}{
			\begin{tabular}{c|ccccc}
				\toprule%
				\multicolumn{6}{c}{P3M-500-NP} \\
				\midrule
				Method & SAD $\downarrow$  & MSE $\downarrow$  & MAD $\downarrow$  & Grad $\downarrow$  & Conn $\downarrow$ \\
				\midrule
				DIM\cite{dim} & 6.3776 & 0.0010 & 0.0037 & 8.9485 & 6.5622 \\
				AlphaGAN\cite{alphagan} & 6.8403 &	0.0013 & 0.0046 & 17.2361 & 7.1052 \\
				IndexNet\cite{attention2} & 6.9850  & 0.0015 & 0.0047 & 16.8927 & 6.9527 \\
				\midrule
				HATT\cite{qiao2020attention} & 37.4163 & 0.0094 & 0.0214 & 36.0780 & 21.7506 \\
				SHM\cite{shm} & 26.4948 & 0.0120 & 0.0152 & 36.8403 & 14.2694 \\
				MODNet\cite{modnet} & 12.3742 & 0.0040 & 0.0071 & 12.5120 & 11.3966 \\
				Ours & 11.6399 & 0.0035 & 0.0067 & 12.4162 & 11.0511 \\
				\bottomrule
			\end{tabular}
		}
	\end{center}
	\caption{Results of trimap-based methods and trimap-free methods on P3M-500-NP testing dataset.}
	\label{tab2}
	\vspace{-.8em}
\end{table}
\subsection{Ablation Studies}\label{ssec4.3}
We perform ablation study of SGM-Net on dataset P3M-500-P. It can be seen from Table \ref{tab3} that compared with the MODNet, the network with the foreground probability map module can achieve better results.
The results of Ex.{\romannumeral1} (MODNet) and Ex.{\romannumeral2} show that the module we designed can effectively fuse the features of the segmented network, and generate a priori information to assist the detail prediction branch to complete the information extraction, e.g., 9.7328 SAD vs 9.34062 SAD, 9.0085 Conn vs 8.8537 Conn. The results of Ex.{\romannumeral2}  and Ex.{\romannumeral3} (our method) show that the output of the segmented network ($\emph{S}_\emph{p}\emph{(I)}$) will interfere with the branch extracted from the probability prediction map, and affect the actual effect of our module, e.g., 9.34062 SAD vs 9.15526 SAD of ours, 8.8537 Conn vs 8.7354 Conn of ours.
\begin{table}
	\begin{center}
		\scalebox{0.99}{
			\begin{tabular}{ccc | ccccc}
				\toprule%
				& PT-G & $S_p(I)$ & SAD $\downarrow$  & MSE $\downarrow$  & MAD $\downarrow$  & Grad $\downarrow$  & Conn $\downarrow$ \\
				\midrule
				\romannumeral1  & & $\surd$ & 9.7328 & 0.0031 & 0.0056 & 13.4755 & 9.0085   \\
				\romannumeral2  & $\surd$ & $\surd$ & 9.3406 & 0.0029 & 0.0054 & 13.4160 & 8.8537  \\
				\romannumeral3  & $\surd$& &\textbf{9.1552} & \textbf{0.0027} & \textbf{0.0053} & \textbf{13.3744} & \textbf{8.7354}  \\
				\bottomrule
			\end{tabular}
		}
	\end{center}
	\caption{Ablation of SGM-Net.  FP-G: The module of foreground probability map generation.  $\emph{S}_\emph{p}\emph{(I)}$ : Using the output of semantic branch as one of the inputs of the detail prediction branch or not. `$\downarrow$' means lower is better.}
	\label{tab3}
	\vspace{-.8em}
\end{table}
\subsection{Results on Real-World Data}\label{ssec4.4}
\begin{figure}[htbp]
	\centering
	\subfigure[Image]
	{
		\begin{minipage}[b]{0.2\linewidth}
			\centering
			\includegraphics[scale=0.4]{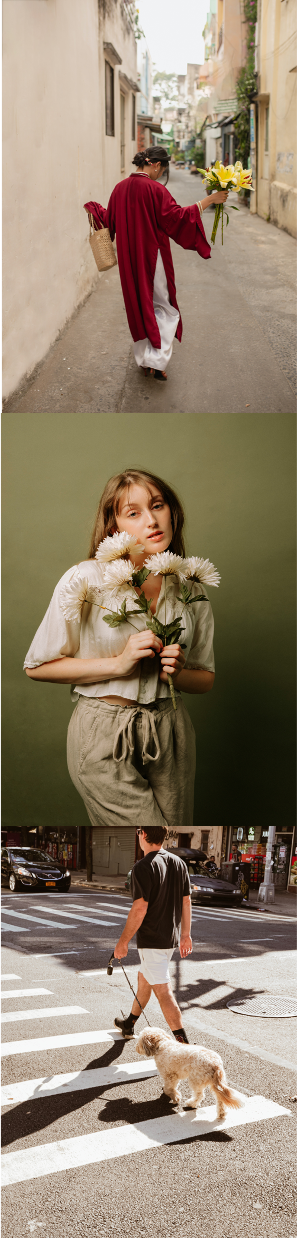} 
		\end{minipage}
	}
	\subfigure[Alpha]
	{
		\begin{minipage}[b]{0.2\linewidth}
			\centering
			\includegraphics[scale=0.4]{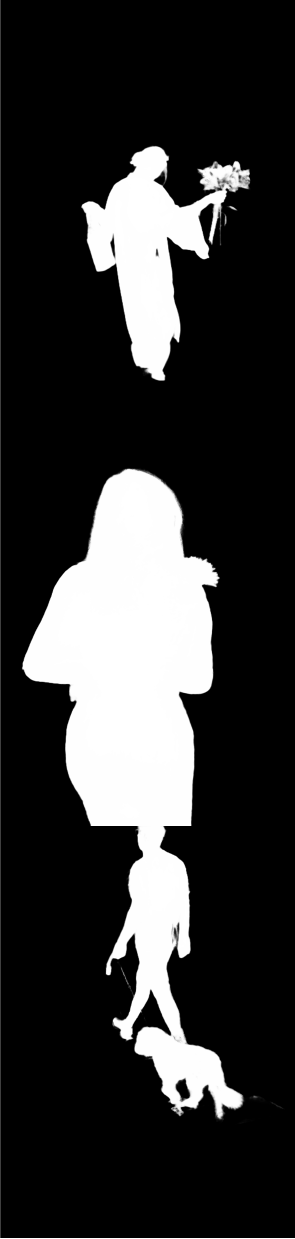} 
		\end{minipage}
	}
	\subfigure[Foreground]
	{
		\begin{minipage}[b]{0.2\linewidth}
			\centering
			\includegraphics[scale=0.4]{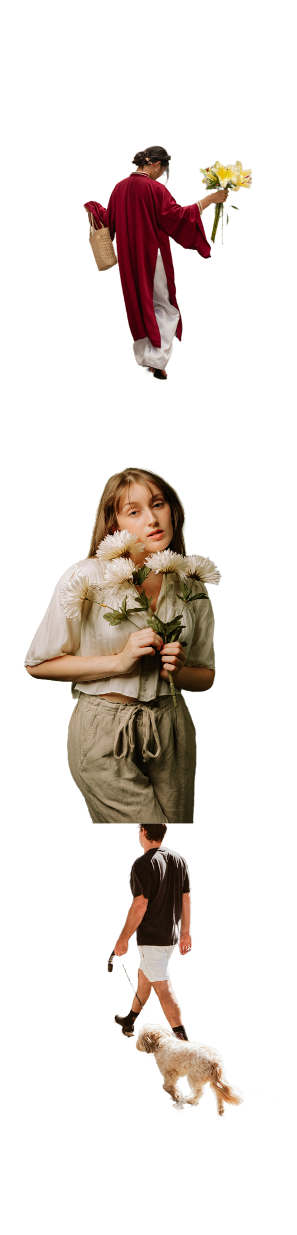} 
		\end{minipage}
	}
	\subfigure[Composition]
	{
		\begin{minipage}[b]{0.2\linewidth}
			\centering
			\includegraphics[scale=0.4]{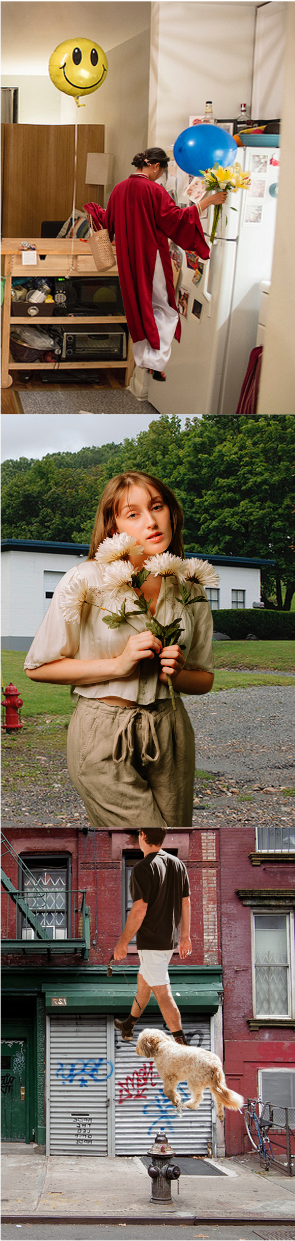} 
		\end{minipage}
	}
	\caption{The results of our method on the real-world data.}\label{fig6}
\end{figure}

In order to study the ability of our model to extend to real-world data, we apply our model to a large number of real-world images for qualitative analysis. Fig. \ref{fig6} shows some visual results. It can be seen that our method still has good performance even in a complex background. Note that the bouquet held by the female in the second image in Fig. \ref{fig6} can also be well separated by our method. In addition, we can also separate the pets far away from the man's body in the third image. These show the good performance of our method for human body extensions (small objects). The last column in Fig. \ref{fig6} shows examples of synthesis with the help of automatic prediction of the foreground and the new background of the alpha matte, which have good visual quality.

\section{Conclusion}\label{sec5}
In this paper, we focus on the problem of human matting. We design a foreground probability map generation module, add it to MODNet, and adjust the whole matting network accordingly to make the transition area smoother, so as to get SGM-Net. The use of green screen in human matting is avoided, and only RGB image is used as input to obtain a high-quality alpha matte. SGM-Net shows good performance on P3M-10k dataset and various real-world data, and is obviously better than MODNet. Although it is not as good as some matting methods based on trimap, the performance gap between them is greatly reduced.


\bibliographystyle{spbasic} 
\bibliography{sn-bibliography}


\end{document}